\documentclass{article}

\usepackage{PRIMEarxiv}

\usepackage[utf8]{inputenc} % allow utf-8 input
\usepackage[T1]{fontenc}    % use 8-bit T1 fonts
\usepackage{hyperref}       % hyperlinks
\usepackage{url}            % simple URL typesetting
\usepackage{booktabs}       % professional-quality tables
\usepackage{amsfonts}       % blackboard math symbols
\usepackage{nicefrac}       % compact symbols for 1/2, etc.
\usepackage{microtype}      % microtypography
\usepackage{lipsum}
\usepackage{fancyhdr}       % header
\usepackage{graphicx}       % graphics

\usepackage{tabularx}
\usepackage{standalone}
\usepackage{gensymb}
\usepackage{hyperref}
\usepackage{mathpazo}
\usepackage{longtable}
\usepackage[normalem]{ulem}
\usepackage{float}
\usepackage{svg}
\usepackage{breakurl}
\usepackage{refcount}
\usepackage{url}
\usepackage[flushleft]{threeparttable}
\DeclareMathAlphabet{\pazocal}{OMS}{zplm}{m}{n}

\usepackage{amsmath,amsthm,amsfonts,dsfont,mathtools} %pacotes matemáticos
\usepackage{multirow}
\usepackage{rotating}
\usepackage{color, colortbl}
\usepackage{xcolor}

\graphicspath{{media/}}     % organize your images and other figures under media/ folder

\title{Neurosymbolic AI and its Taxonomy: a Survey
\thanks{\textit{\underline{Citation}}: 
\textbf{Gibaut et al. Neurosymbolic AI and its Taxonomy: a Survey DOI:10.48550/arXiv.2305.08876.}} 
}

\author{
  Wandemberg Gibaut\\
  Eldorado Institute of Technology\\
  Campinas, S\~{a}o Paulo\\
  \texttt{wandemberg.gibaut@eldorado.org.br} \\
  \And
  Leonardo Pereira \\
  Eldorado Institute of Technology\\
  Campinas, S\~{a}o Paulo\\
  \texttt{leonardo.pereira@eldorado.org.br} \\
  \And
  Fabio Grassiotto \\
  Eldorado Institute of Technology\\
  Campinas, S\~{a}o Paulo\\
  \texttt{fabio.grassiotto@eldorado.org.br} \\
  \And
  Alexandre Osorio \\
  Eldorado Institute of Technology\\
  Campinas, S\~{a}o Paulo\\
  \texttt{lalexandre.osorio@eldorado.org.br} \\
  \And
  Eder Gadioli \\
  Eldorado Institute of Technology\\
  Campinas, S\~{a}o Paulo\\
  \texttt{eder.gadioli@eldorado.org.br} \\
  \And
  Amparo Munoz \\
  Eldorado Institute of Technology\\
  Campinas, S\~{a}o Paulo\\
  \texttt{amparo.munoz@eldorado.org.br} \\
  \And
  Sildolfo Gomes \\
  Eldorado Institute of Technology\\
  Campinas, S\~{a}o Paulo\\
  \texttt{sildolfo.gomes@eldorado.org.br} \\
  \And
  Claudio Filipi Goncalves do Santos \\
  Eldorado Institute of Technology\\
  Campinas, S\~{a}o Paulo\\
  \texttt{claudio.santos@eldorado.org.br} \\
}

\begin{document}
\maketitle

\begin{abstract}
  Neurosymbolic AI deals with models that combine symbolic processing, like classic AI, and neural networks, as it's a very established area. These models are emerging as an effort toward Artificial General Intelligence (AGI) by both exploring an alternative to just increasing datasets' and models' sizes and combining Learning over the data distribution, Reasoning on prior and learned knowledge, and by symbiotically using them.
  This survey investigates research papers in this area during the recent years and brings classification and comparison between the presented models as well as applications.
\end{abstract}

% keywords can be removed

%\documentclass[acmsmall]{acmart}
%\begin{document}

%\newpage
\section{Introduction}
\label{s.introduction}
%\rowcolors{2}{gray!25}{white}

As Artificial Intelligence, and Deep Learning in particular, reach impressive results, it gains also unprecedented popularity not only in academics and industry but also in popular culture and society in general. This increasingly ubiquitous AI presence has arisen several concerns about its impacts on humanity and the planet, with some well-known scientists like Stephen Hawking having spoken concerns about AI's accountability \cite{russell2015ethics}.

Despite achieving outstanding results in Computer Vision, Natural Language Processing and Game Playing \cite{lecun2015deep, silver2017mastering}, tasks in which AIs formerly have poor performance compared to humans, those concerns about AI triggered debates among research communities, including those discussed by Gary Marcus \cite{marcus2020next} and on AAAI-2020 debate with Geoffrey Hinton, Yoshua Bengio and Yann LeCun \cite{kahneman2020aaai20}.

These discussions have a common ground about the future of AI: the aim during the next decades is to investigate and build richer AI systems that are explainable, trustworthy, and based on solid principles that unify the ability to learn from experience and to reason from what has been learned \cite{valiant2003three}. As discussed by Daniel Kahneman\cite{kahneman2011thinking, kahneman2020aaai20}, there is a need for a "System 2" for Language (and symbol) manipulation.

Here, Neurosymbolic Computing appears as a strong candidate to fill those gaps by integrating learning from the environment in a connectionist fashion and reasoning from what has been learned using symbolic processing and representation. Table \ref{tab:methods} shows a summary of the surveyed methods that are talked about in this survey while Table \ref{tab:papers} shows the papers we analyzed, also with a brief description.

\begin{table}[h!]
    \centering
    \caption{Brief explanation of surveyed methods}
    \label{tab:methods}
    \resizebox{\columnwidth}{!}{%
  \begin{tabular}{@{}ll@{}}
  \toprule
\rowcolor{gray}
  Method &
    Description \\ \midrule
  Logic Tensor Networks (LTN)~\cite{Badreddine_2022} &
  \begin{tabular}[c]{@{}l@{}}LTN is a Neurosymbolic computational model and \\ formalism. It supports learning and reasoning through \\ Real Logic - a many-valued, end-to-end differentiable \\ first-order logic - as a representation language for \\ deep learning~\cite{Badreddine_2022}.\end{tabular} \\ \midrule
  Neural Logic Machine (NLM)~\cite{dong2019neural} &
  \begin{tabular}[c]{@{}l@{}}NLM is a Neurosymbolic architecture for inductive\\ learning and logic reasoning. By using neural networks\\ and a symbolic processor~\cite{dong2019neural}.\end{tabular} \\ \midrule
  Logical Boltzmann Machines (LBM)~\cite{tran2021logical} &
  \begin{tabular}[c]{@{}l@{}}LBMs is a Neurosymbolic system capable of representing \\ any propositional logic formulae in strict disjunctive normal\\ form~\cite{tran2021logical}.\end{tabular} \\ \midrule
  Logic Neural Networks (LNN)~\cite{LNN} &
    \begin{tabular}[c]{@{}l@{}}LNN is a novel framework that provides neural networks for\\ learning and symbolic logic for knowledge and reasoning. \\ Every neuron has a meaning as a component in a weighted \\ real-valued logic formulae~\cite{LNN}.\end{tabular} \\ \midrule
  Neuro-Symbolic Concept Learner (NS-CL)~\cite{CLEVRER} &
  \begin{tabular}[c]{@{}l@{}}NS-CL is a model that is capable of learning visual concepts,\\ words, and semantic parsing of sentences without explicit \\ supervision. The model learns by "looking" at images and \\ reading paired questions and answers, thus building an \\ object-based scene representation and translating sentences \\ into symbolic programs that can be executed~\cite{CLEVRER}.\end{tabular} \\ \midrule
  Generative Neurosymbolic Machines (GNM)~\cite{jiang2020generative} &
  \begin{tabular}[c]{@{}l@{}}GNMs is a generative model that combines the benefits of distributed\\ and symbolic representations to support both structured\\ representations of symbolic components and\\ density-based generation~\cite{jiang2020generative}.\end{tabular} \\ \bottomrule
  \end{tabular}
    }
    \end{table}
%\begin{tabular}[c]{@{}l@{}}GNNs are a neural model that can capture the dependence \\ of graphs using message passing between the nodes of graphs~\cite{GNN}.\end{tabular} \\ \bottomrule

\small
\begin{table}[h!]
  %\centering
    \caption{Surveyed papers and their descriptions}
    \label{tab:papers}
    \resizebox{\columnwidth}{!}{%
    \begin{tabular}{cl}
  \hline
\rowcolor{gray}
  Paper Name                                                                                                                                           & Description                                                                                                                                                                                                        \\ \hline
  Logic Tensor Networks                                                                                                                                & \begin{tabular}[c]{@{}l@{}}"...a neurosymbolic framework that\\ supports querying, learning and \\ reasoning with both rich data and \\ abstract knowledge about\\ the world."~\cite{Badreddine_2022}\end{tabular}                        \\\midrule
  Neural-Symbolic Integration for Fairness in AI                                                                                                       & An application of LTN for AI Fairness                                                                                                                                                                              \\
  \begin{tabular}[c]{@{}c@{}}Logic Tensor Networks for \\ Semantic Image Interpretation\end{tabular}                                                   & \begin{tabular}[c]{@{}l@{}}An application of LTN for \\ Semantic Image Interpretation\end{tabular}                                                                                                                 \\\midrule
  Neurosymbolic AI for Situated Language Understanding                                                                                                 & \begin{tabular}[c]{@{}l@{}}An application of LTN with multimodal data\\ fo Language Understanding\end{tabular}                                                                                                     \\\midrule
  \begin{tabular}[c]{@{}c@{}}Neural-Symbolic Reasoning \\ Under Open-World and Closed-World Assumptions\end{tabular}                                   & \begin{tabular}[c]{@{}l@{}}A study comparing LTN with purely \\ symbolix reasoning\end{tabular}                                                                                                                    \\\midrule
  Neural Logic Machines                                                                                                                                & \begin{tabular}[c]{@{}l@{}}"... a neural-symbolic architecture for both \\ inductive learning and logic reasoning."~\cite{dong2019neural}\end{tabular}                                                                                   \\\midrule
  Logical Neural Networks                                                                                                                              & \begin{tabular}[c]{@{}l@{}}"... a novel framework seamlessly providing \\ key properties of both neural nets (learning)\\  and symbolic logic \\ (knowledge and reasoning)."~\cite{LNN}\end{tabular}                          \\\midrule
  \begin{tabular}[c]{@{}c@{}}LOA: Logical Optimal Actions for \\ Text-based Interaction Games\end{tabular}                     & \begin{tabular}[c]{@{}l@{}}An application of LNN to Language \\ Interaction Game with Reinforcement \\ Learning\end{tabular}                                                                                       \\\midrule
  \begin{tabular}[c]{@{}c@{}}LNN-EL: A neuro-symbolic approach \\ to short-text entity linking.\end{tabular}                                           & \begin{tabular}[c]{@{}l@{}}An application of LNN to Natural Language\\ Processing\end{tabular}                                                                                                                     \\\midrule
  \begin{tabular}[c]{@{}c@{}}Leveraging abstract  meaning representation \\ for knowledge base question answering\end{tabular}                         & \begin{tabular}[c]{@{}l@{}}"... a modular KBQA system, that \\ leverages (1) Abstract Meaning \\ Representation (AMR) parses for \\ task-independent question \\ understanding ..."~\cite{KBQA}\end{tabular}                   \\\midrule
  Logical Boltzmann Machines                                                                                                                           & \begin{tabular}[c]{@{}l@{}}"...  a neurosymbolic system that can \\ represent any propositional logic \\ formula in strict disjunctive normal form."~\cite{tran2021logical}\end{tabular}                                                  \\\midrule
  \begin{tabular}[c]{@{}c@{}}Expert Knowledge Induced Logic Tensor Networks: \\ A Bearing Fault Diagnosis Case Study\end{tabular}                      & \begin{tabular}[c]{@{}l@{}}An application of LTN to bearing fault \\ diagnosis\end{tabular}                                                                                                                        \\\midrule
  Generative Neurosymbolic Machines                                                                                                                    & \begin{tabular}[c]{@{}l@{}}A generative model that combines symbolic\\ and neural processing to  support both \\ structured representations of \\ symbolic components and \\ density-based generation\end{tabular} \\\midrule
  \begin{tabular}[c]{@{}c@{}}The Neuro-Symbolic Concept Learner: \\ Interpreting Scenes, Words, and Sentences \\ From Natural Supervision\end{tabular} & \begin{tabular}[c]{@{}l@{}}"... a model that learns visual concepts, \\ words, and semantic parsing of sentences \\ without explicit supervision on any of \\ them; instead..."~\cite{CLEVRER}\end{tabular}\\ \bottomrule               
  \end{tabular}
    }
  \end{table}

A brief intermission by defining what symbols are is a good place to start, with the following subsection.

\subsection{Symbols}

In the classical article `Computer science as empirical inquiry: Symbols and search', Alan Newell and Herbert Simon introduced the concept of \emph{Physical Symbol Systems} as the required building blocks for intelligent systems. Symbolic AI had the initial intent of modeling specialist systems through the processing of symbolic inputs in what is known as inference engines \cite{newell2007computer}.

In the article, it is hypothesized that symbols used within a computational system are the very same representations humans use for daily life, with the logical implication that we humans are instances of physical symbol systems. This concept, familiar to anyone involved with Computing Science, comes from the formalization of logic.

The \emph{Physical Symbol Hyphothesis} states, then, that for a physical system to show general intelligence action, it must be a physical symbol system. The hypothesis can be explained by three main pillars: Necessity, Sufficiency, and General Intelligent Action.

\emph{Necessity} can be explained in the sense that any physical system that exhibits general intelligence will be one implementation of a physical symbol system.

\emph{Sufficiency} in the sense that all physical symbol systems can be further organized to support general intelligence.

\emph{General Intelligent Action} is supposed to be the scope of intelligence as can be seen in humans, i.e., the behavior that is capable of supporting adaptation to an environment.

\subsection{Methodoly and Scope of this Work}

The objective of this survey is to provide an overview of the state-of-the-art of Neurosymbolic AI, which seeks to combine the reasoning and explainability of symbolic representations of network models with the robust learning that can be achieved with neural networks. Also, this paper focuses on works published from 2019 until now. Previous surveys can also be found \cite{garcez2019neural, de2020statistical, lamb2020graph}. Compared to~\cite{garcez2019neural} our work brings papers published since 2019, while their work was published in 2019 analysing works prior to that, our work is also focus on applications of neurosymbolic system which they did not. The work made in~\cite{de2020statistical} limits itself to a logical and probabilistic perspective, not focusing in frameworks or applications. In~\cite{lamb2020graph} their focus was mainly in neuro-symbolic AI and graph combination and, while a few frameworks analysed overlap with our analysis they did not metioned Neural Logic Machine (NLM)~\cite{dong2019neural}, Logical Boltzmann Machines (LBM)~\cite{tran2021logical}, Logic Neural Networks (LNN)~\cite{LNN} or Generative Neurosymbolic Machines (GNM)~\cite{jiang2020generative}. Furthermore our survey is the only survey among them that ties everything together, such as logic, frameworks, applications, etc...

The rest of this work is divided as follows: Section \ref{s.knowledge} discusses how each of the models we analyze in this paper deals with the Knowledge Representations in their systems. Section \ref{s.learning} concerns a core topic on Neurosymbolic AI: how each system implements its learning processes and its relations with pure Symbolic and pure connectionist systems. Section \ref{s.reasoning} shows how Neurosymbolic systems reason about their knowledge and data extrapolation. Section \ref{s.explain} discusses Explainability and Trustworthiness in Neurosymbolic systems, while Section \ref{s.applications} presents some interesting applications for those systems. Finally, Section \ref{s.conclusion} wraps everything we discussed, presenting some conclusions and a table where one can visualize how each model is located in the presented taxonomy. In this here survey we will be using existing taxonomy in the field.

%\end{document}

%\orcid{0000-0002-8717-6097}

%\documentclass[acmsmall]{acmart}
%\begin{document}

\section{Knowledge Representation in Neurosymbolic Systems}
\label{s.knowledge}

This section discusses how knowledge is represented in neurosymbolic systems.~\cite{learning&reasoning} argues that logical calculus can be carried out exactly or approximately by a neural network, so the knowledge representation would be the basis of a neural-symbolic system that provides a mapping mechanism between symbolism and connectionism. On one hand, a symbolic system knowledge is commonly represented as a discrete set of symbols~\cite{levesque1986knowledge}. Take for example a multi-class classification problem where we have four classes. These four classes are represented either as binary values or as a 4-state variable. On the other hand, neuron-like systems often adopt continuous output space as a standard.  This can be seen in deep learning systems such as regression tasks.

%\subsection{Tensorization}
Both the LTN\cite{Badreddine_2022} and the LNN\cite{LNN} frameworks analyzed in this section use First Order Logic (FOL). FOL enables the representation of a certain sphere in terms of objects that already exist in that sphere and relations that hold between certain objects in that sphere. As FOL can be treated as a language its vocabulary includes constant symbols, predicate symbols, and functional symbols. In this domain, constants refer to objects, while predicates and functions refer to relations.

First, it is analyzed how Logic Tensor Networks (LTN) \cite{Badreddine_2022} deal with knowledge. The concept behind LTN's vision of logic is Real Logic. Real Logic, simply put, is a way of representing symbolic knowledge so that it can be applied in a Machine Learning environment \cite{real_logic}. Its main goal is to achieve a vector-based representation technique which should be shown adequately for integrating machine learning and symbolic reasoning in a grounded way. To achieve this they use fuzzy semantics (the truth value of a logical formula is between $0$ and $1$ instead of being just true or false).

The first step to this system is a language $\mathcal{L}$, which in this case would be First Order Logic. Once the language has been established we need the constant symbols $\mathcal{C}$ to write with a given language, an example of these symbols would be $\forall$, $\notin$ , etc... It also needs functional symbols $\mathcal{F}$ such as $\lnot $ or $\rightarrow$. Finally,  predicate symbols $\mathcal{P}$ should be used, which are an amalgamation of the two, meaning a set of relational symbols.

The semantics of this language in the "real world" comes in the form of groundings $\mathcal{G}$. Groundings are the way Real Logic maps the language $\mathcal{L}$ into the "real world" with real numbers in the form of n-tuples. Grounding also means that domains $\mathcal{D}$ are interpreted as tensors in the "real world" if one so desires. To ground an atomic formula quantifiers and connectives are needed, those being $\lnot $, $\rightarrow$ and $\wedge$. Attention is needed to the fact that the term grounding does not take this meaning in other contexts but in Real Logic the term is used to drive home the point that these semantics are grounded in real values.

For a better grasp of this concept, an example is presented. Given a language $\mathcal{L}$, constant symbol $\mathcal{C}$, $\forall$, and functional symbols $\mathcal{F}$, $\rightarrow$. We can write a predicate $\mathcal{P}$ such as:\linebreak $\forall$ x $\rightarrow$ Class(Positive).. With this equation, we now ground these symbols with our grounding $\mathcal{G}$ for a real value representation. This real representation could be a tensor which would represent the degree of truth of this statement in our dataset or neural network.

Logic Neural Networks (LNN)~\cite{LNN} also use Real-Valued Logic when representing symbolic knowledge, but instead of static values, it uses truth bounds. LNN also can not use functions or equality symbols and, since equality symbols are not handled, it makes the the unique-names assumption, predicating that every constant symbol refers to a distinct object in the domain. One distinct aspect of LNNs is that they introduce two quantifiers when dealing with FOL, those being \emph{universal} and \textit{existential}.

In LNNs there is an atomic neuron for each predicate symbol, a neuron for logical connectivity, and a connective neuron. Each neuron carries a table whose columns are sorted by unique variables appearing in the represented sub-formulae, and rows sorted by a set of n-element tuples of groundings, where the tuple size n corresponds to the arity of the neuron. The contents of the tables are the bounds - as stated before - of each grounding when substituted for the variables.

When dealing with bounds they can be, if well explored, more interesting than static values: "The bounds of a universal quantifier node are set to the bounds of the grounding with the lowest upper bound, so that if all of its groundings are True then the universal statement is also True. And the bounds of the existential quantifier are set to the bounds of the grounding with the highest lower bound so that if this lower bound is True the existential statement will be True when at least one of its groundings is True" \cite{LNN}. Dealing with LNNs also allows the exploration of the concept of inference for First Order Logic. When exploring inference for FOL all neurons in the LNN return tables of groundings and their corresponding bounds. Neural activation functions are then mutated to perform agglutination over columns that share variables while still computing truth value bounds associated with certain groundings. Inverse activation functions are modified similarly but are then filtered, reducing results over non-significant variables so that we can have the tightest bounds.

\cite{wagner2022neural} explain that LTNs are instances of the Differentiable Fuzzy Logic (DFL) method, which employs fuzzy logic with differentiable logical operators. Loss functions that maximize the satisfiability of the symbolic knowledge base are built into this method. Since the fuzzy logical operators are differentiable, well-known methods like gradient descent can be employed in this optimization.

Neural Logic Machines (NLM) are presented by~\cite{dong2019neural} as Neurosymbolic models designed to address some challenges that, allegedly, pure connectionist Neural Network models cannot address. According to the authors, those challenges are:

\begin{itemize}
    \item The learning system should recover a set of lifted rules (i.e., rules that apply to objects uniformly instead of being tied with specific ones) and generalize to scenarios with different data than the presented.
    \item The learning system should deal with high-order relational data and quantifiers, which go beyond the scope of typical graph-structured neural networks like, jointly inspecting a set of objects to apply transitivity rules.
    \item The learning system should scale up regarding the complexity of the rules. The existing logic-driven approaches suffer an exponential computational complexity as the number of logic rules to be learned increases (also known as ”the Curse of Dimensionality”).
    \item The learning system should recover rules based on a minimal set of learning priors.
\end{itemize}

Concerning the Knowledge Representation in NLMs, the model adopts probabilistic tensor representation for the logic predicates, that are grounded in a set of objects. Also, NLMs have a ” breath’ property that is intrinsically related to the concept of arity, which represents how many objects are considered in the logical rule (i.e. if a rule is about an object itself, about a relationship between two of them, and so on). The total ”width” of the model will also take into account the number of prepositions to be learned. Figure \label{fig:nlm} illustrates this.

\begin{figure}
    \centering
    \includegraphics[width=1.0\columnwidth]{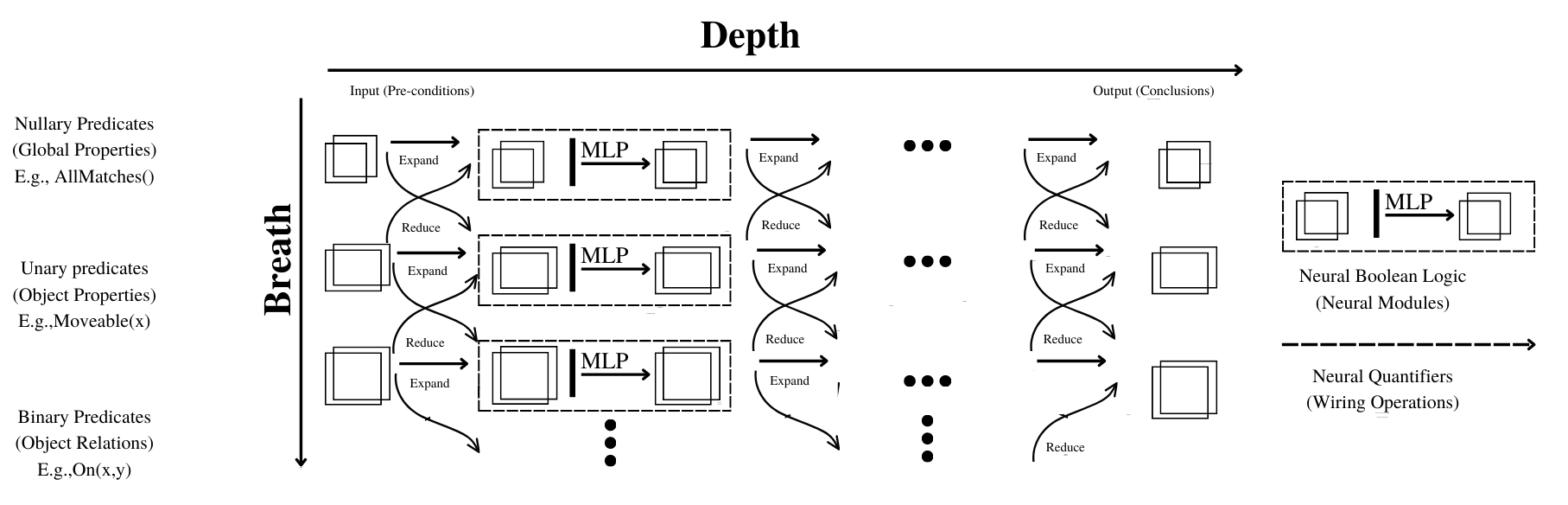}
    \caption{ An illustration of Neural Logic Machines (NLM). During forward propagation, NLM takes object properties and relations as input, performs sequential logic deduction, and outputs conclusive properties or relations of the objects. Adapted from \cite{dong2019neural}}
    \label{fig:nlm}
\end{figure}

\begin{figure}[b]
    \centering
    \includegraphics[width=0.8\columnwidth]{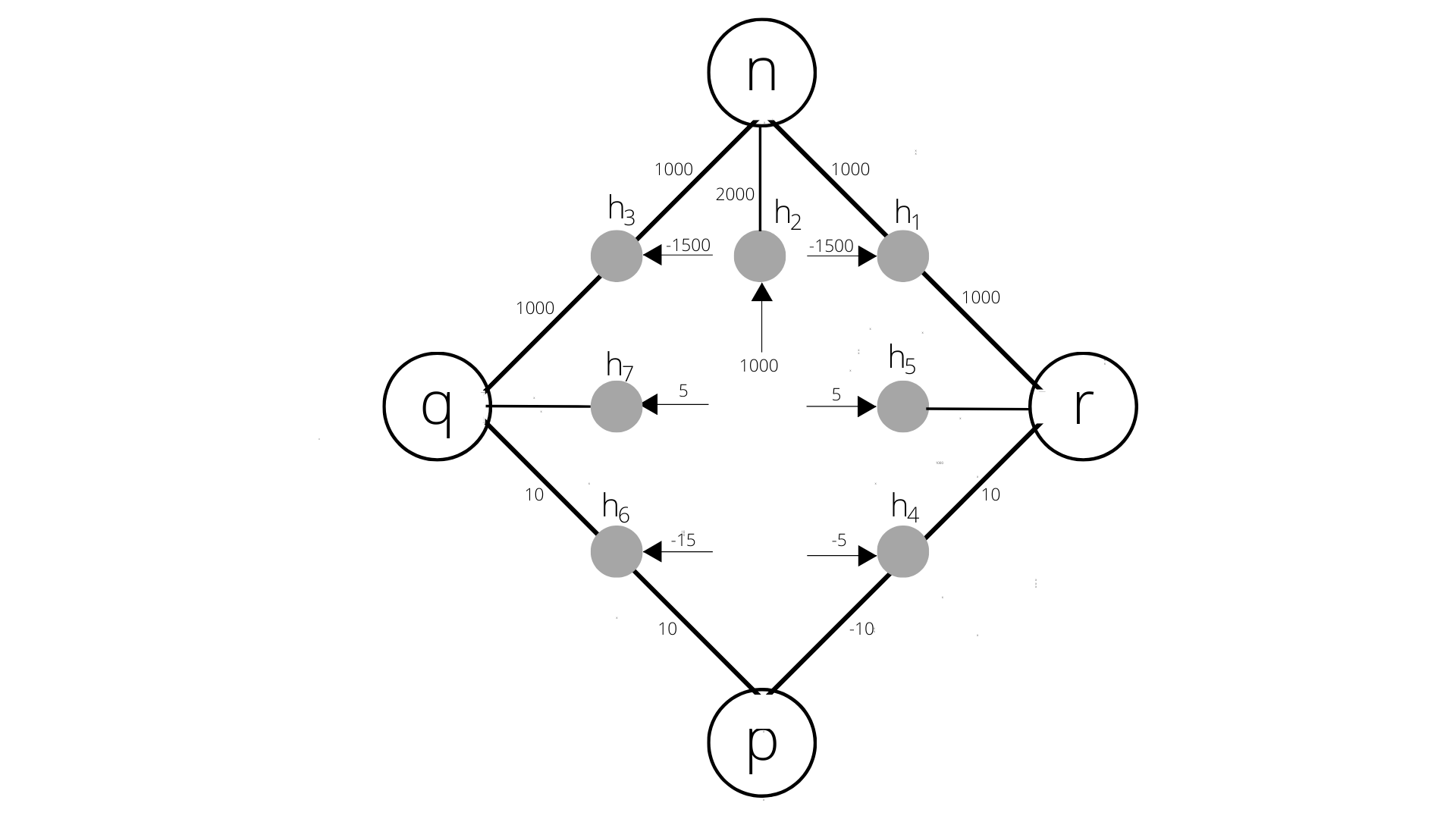}
    \caption{ An example of RBM for the Nixon diamond problem. With  = 0.5, this RBM has energy function: E = -h1(1000n+1000r-1500)-h2(-2000n+1000)-h3(1000n+1000q-1500)-h4(10r-10p-5)-h5(-10r+5)-h6(10q+10p-15)-h7(-10q+5). Adapted from \cite{tran2021logical}}
    \label{fig:lbm}
\end{figure}

Logical Boltzmann Machines (LBM) appear first in~\cite{tran2021logical} as a Neurosymbolic system that can represent any propositional logic formula in strict disjunctive form. The system relies on the equivalence between the logical satisfiability of formulas and the energy minimization in Restricted Boltzmann Machines. An LBM system is built upon a weighted knowledge base, with its topology reflecting that prior knowledge. Here, each propositional logic formula is mapped into a Strict Disjunctive Normal Form (SDNF), that is, a disjunction of conjunctions with at most one conjunctive clause that maps to True for any assignment of truth-values x. An example of LBM for the Nixon diamond problem is shown in Figure~\ref{fig:lbm}. The weighted knowledge base used to construct this network was:

\begin{itemize}
    \item 1000 : n → r \textit{Nixon is a Republican}.
    \item 1000 : n → q \textit{Nixon is also a Quaker}.
    \item 10 : r → ¬p \textit{Republicans tend not to be Pacifists}.
    \item 10 : q → p \textit{Quakers tend to be Pacifists}.
\end{itemize}

Another Neurosymbolic system we will explore is the NeuroSymbolic Concept Learner (NS-CL) presented by \cite{CLEVRER}. This model consists of an object-based scene representation and sentence translation into executable symbolic programs. They use neurosymbolic reasoning to execute these programs on the latent scene representation. The logic behind the project is to emulate human concept learning by allowing the perception module to learn visual concepts based on the language description of the object referred to.

Just as humans are capable of learning visual concepts by jointly understanding vision and a given language, the authors propose the NS-CL a system that jointly learns visual perception, words, and a semantic language parsing from images and pairs of question-answer. The NS-CL consists of three modules those being: a neural-based perception module (this module is used to extract object-level representations from a given scene), a visually- grounded semantic parser (this module translates questions into executable programs) and a symbolic program executor (the module that will read out the perceptual representation of objects, classifies their attributes, and relations, and executes the program to obtain an answer).

The NS-CL learns from natural supervision, for example, images and question-answer pairs. This means that it requires no annotations on images or semantic programs for sentences. Just as a human would, this system learns via curriculum learning. NS-CL will start by learning concepts or representations of individual objects from short questions on simple scenes, for example,’ What color is the cube?’. The final result is learning object-based concepts such as colors and shapes. The next step is learning relational concepts by leveraging object-based concepts to interpret object referrals, for example,’ Is there a sphere next to the cube?’. By default, the model will iteratively adapt to more complex scenes and highly compositional questions. The authors argue that NS-CL’s modularized designenables interpretable, robust, and accurate visual reasoning. As evidence, they claim that it achieves state-of-the-art performance on the CLEVR dataset \cite{CLEVR}. %Figure \ref{fig:ns-cl} ilustrates how this framework works.

%\begin{figure}
%    \centering
%    \includegraphics[width=1.0\columnwidth]{media/NS-CL.png}
%    \caption{Neuro-Symbolic Concept Learner execution flowchart}
%    \label{fig:ns-cl}
%\end{figure}

%\end{document}
%\orcid{0000-0002-8717-6097}

%\documentclass[acmsmall]{acmart}
%\begin{document}

\section{Learning in Neurosymbolic Systems}
\label{s.learning}

There are some conceptually different learning processes in neurosymbolic systems regarding the integration of prior knowledge and data generalization. \cite{learning&reasoning} lists methods that approach learning and neurosymbolic systems that were tested and proved. This section explains a few of those methods.

\subsection{Inductive Logic Programming}

%/* Esse conceito já não foi apresentado no parágrafo anterior? */ Inductive logic programming (ILP) can take advantage of the learning capability of neurosymbolic computing to automatically construct a logic program from examples. Normally, approaches in ILP are categorized into bottom-up and top-down.
%Bottom-up approaches construct logic programs by extracting specific clauses from examples and, after that, proceeds to search for more general clauses. Top-down approaches, on the other hand, construct logic programs from the most general clauses and extend them to be more specific.

First we have Inductive Logic Programming (ILP). In an ILP system, we take advantage of the learning capability of neurosymbolic computing to generate a logic program automatically from examples. We can approach ILP as bottom-up or top-down. In a bottom-up approach, a logic program is built by extracting specific clauses from examples and, after that, generalization procedures are applied in search of more general clauses~\cite{learning&reasoning}. The disadvantage of this approach is that it highly depends on how well the generalization can be. In the field of neural-symbolic computing, bottom clause propositionalisation is a popular approach because bottom clause literals can be encoded into the neural network directly as data features while presenting their semantic meaning.

On the other hand, in a top-down approach, a logic program is built by extracting general clauses from examples. The resulting program is then extended to specific problems~\cite{learning&reasoning}. The most popular idea concerning this approach is to take advantage of neural networks' inference and learning capabilities to fine-tune and test the quality of our rules by replacing logical operations with differentiable operations. Starting with the most general clauses, a top-down based system creates rules from facts. We can cite Neural Logic Programming (NLP)~\cite{yang2017differentiable} as an example. In NLP learning of rules are based on the differentiable inference of TensorLogs. In this application matrix computations are used to soften logic operators where confidence of conjuctions and confidence of disjunctions are computed as product and sum, respectively.

In Neural Logic Machines (NLM) \cite{dong2019neural}, the learning process is given by inputting tensors representing predicates of different \emph{arities} (''facts'', considering the Closed-World Assumption~\cite{minker1982indefinite}) from a knowledge base and the model outputs tensors representing new predicates. For example, in the Family Tree example presented by the authors, the input predicates are \emph{IsSon}, \emph{IsDaughter}, \emph{IsFather} and \emph{IsMother} and the outputs (conclusions) are \emph{HasFather}, \emph{HasSister}, \emph{IsGrandparent}, \emph{IsUncle} and \emph{IsMGUncle}.

\subsection{Hybrid Learning}
%\subsection{Vertical Hybrid Learning}
%?? allegedly Vertical, but it seens outdated and modified in the new version

Another important approach to learning in a neurosymbolic context is Horizontal Hybrid Learning (HHL), which is a neurosymbolic answer to data problems. Techniques such as deep learning usually require large amounts of data to find statistical regularities. However, the access to large datasets can be difficult and, as we know, a small dataset can make models susceptible to overfitting. On the other hand, a neurosymbolic system has the advantage of generalization if prior knowledge is provided. This characteristic is achieved by combining logical formulas with data during the learning process while using data to fine-tune our knowledge at the same time. We can improve the efficiency and effectiveness of a neural network by encoding logical knowledge as controlled parameters while training models~\cite{effectiveness,learning&reasoning}. When lacking prior knowledge the ideia of neural-symbolic integration for knowledge transfer learning can be applied. This solution aims to extract symbolic knowledge from a related domain and transfer it to improve the learning in another domain, starting from a network that does not need to be instilled with background knowledge.

The next learning approach is Vertical Hybrid Learning (VHL)~\cite{learning&reasoning}, which employs an idea similar to how human brains work. Neuroscientific studies show that different areas of the brain process different input signals such as logical thinking and emotion processing~\cite{neuroscience}. In VHL we place a logic network on top of a deep neural network to learn what is happening in a black-box system that has a high-level abstraction from complex inputs such as audio, images and text. As an example in~\cite{SemanticImageInterpretation} a Fast-RCNN~\cite{girshick2015fast} is used for bounding-box detection of parts of objects in combination with a Logic Tensor Network, which is used to reason about relations between parts of objects and types of objects, meaning that the perception part (Fast-RCNN) is fixed and learning is carried out in the reasoning part (LTN).

In LTN \cite{Badreddine_2022}, learning refers to the satisfiability of a triple $ T = \langle K,G(\cdot|\theta),\Theta \rangle $ where K is a set of closed first-order logic formulas $G(\cdot|\theta)$ is a parametric grounding for all the symbols $S = D \cup X \cup C \cup F \cup P $ and all the logical operators; and $\Theta = \{\Theta_s\}$ for $s\in S$ is the hypothesis space for each set of parameters $\theta_s$ associated with symbol s. The learning processes may occur on the grounding of constants (embedding), grounding of functions (regression or generative task) or grounding of predicates (classification). The idea in such systems is to embed prior knowledge in the network as axioms.

Although LBM systems~\cite{tran2021logical} are necessarily built using a weighted knowledge base, they can further improve their performance in certain tasks by learning over data. This process is pretty straightforward with train and validation separation and the target variable left out. The inference is performed using a conditional distribution on the target variable.

%\begin{figure}
%    \centering
%    \includegraphics[width=1.0\columnwidth]{media/NS-CL_learning.png}
%    \caption{Neuro-Symbolic Concept Learner learning process. A demonstrates the curriculum learning of visual concepts, words, etc... While B Illustrates theirs neurosymbolic inference model for Visual Question Answering.}
%    \label{fig:ns-cl_learning}
%\end{figure}

The NS-CL learning process consists of its semantic parser generating the hierarchies of latent programs in a sequence to tree manager~\cite{NS-CL_ref1}. They use a bidirectional GRU \cite{NS-CL_ref2} to encode an input question which outputs a fixed-lenght embedding of the question. After that a decoder based on GRU cells is applied to the embedding and recovers the hierarchy of operations as the latent program. 
    
Given the latent program recovered from the question in natural language a symbolic program executor is enabled and derives the answer based on the object-based visual representation. This program executor is a collection of deterministic functional modules which were designed to execute all logic operations specified in the domain-specific language (DSL). 

To make the execution differentiable from the visual representations, the authors represent the intermediate result in a probabilistic manner. A set of objects is represented by a vector, just as the attention mask over all objects in the scene. Each element $Mask_i$ denotes the probability of the n-th object of the scene belonging to the set. The first Filter operation will output a mask with size 4, since there are 4 objects in the scene. Each element will represent the probability that the corresponding object is selected. The output mask on the object will be then fed into the next module as input and the execution continues normally.

The NS-CL training process is split into four stages: learning object-level visual concepts, learning relational questions, learning more complex questions with perception modules fixed and joint fine-tuning of all modules. The authors found out that the joint fine-tuning is essential to their neurosymbolic concept learner.

%\end{document}
%\orcid{0000-0002-8717-6097}

%\documentclass[acmsmall]{acmart}
%\begin{document}

\section{Reasoning in Neurosymbolic Systems}
\label{s.reasoning}

From a purely mathematical perspective, Reasoning is the task of verifying if a certain conclusion is a logical consequence of a set of premises, both interpreted as symbols~\cite{Badreddine_2022}. Although there is other ways, a model-based approach is usually the focus of Neurosymbolic systems, as there is a need to integrate Learning and Reasoning. 

This section explores how reasoning happens in a neurosymbolic context. Reasoning being such an important feature in a neural network system, various approaches have been studied and tested in this context. These approaches can be model-based or theorem proving. Given that in a neurosymbolic enviroment the main focus lies in the integration of reasoning and learning the preferred approach is model-based.

\subsection{Forward and Backward chaining}

The first approach to model-based reasoning is forward and backward chaining. This approach consists of two popular inference techniques for logic programs and systems. For the specific case - this being neurosymbolic systems - both forward and backward chaining are implemented by feedforward inference. As implied by their names, forward and backward chaining are opposites. In a forward chaining context new facts are generated from the head literals of the rules using know facts that exist in the knowledge base. For example, imagine a dataset where we have famous individuals and their contribution to society, we know that Machado de Assis was a writer and that implies he can read so it infers that Frank Herbert could read because he also was a writer. 
    
In contrast, backward chaining does a backward search from a goal in the knowledge base in order to determine if a certain query is possible. Continuing the example given above, in the context of a backward chaining we query if Frank Herbert could read and than find in the knowledge base if another person with the same profession as him could also read. It is important to acknowledge that backward reasoning is expressively harder to implement than forward reasoning.

Inside this classification, NLM~\cite{dong2019neural} can be considered  systems as a model that performs reasoning by forward chaining, since it presents the inputs and outputs of neural networks as grounding tensors of predicates for existing facts and new facts respectively.

\subsection{Approximate Satisfiability}

The second approach to model-based reasoning consists of approximate satisfability. When dealing with arbitrary formulas, inference can be more complex, given that a way to traverse this is to search over our hypothesis space for a solution that maximizes satisfability in the formulas and facts of knowledge base. 
    
Reasoning with maximum satisfability is NP-hard and for this reason neurosymbolic systems offer approximate satisfability. These systems are trained to approximate the best possible satisfability so inference is efficient with feedforward propagation. This approach can be thought as a method that travels in the hypothesis space searching for the point where the restrictions (formulas and facts) are in their most interesting form.

In LTN~\cite{Badreddine_2022}, the reasoning process is given by an approximate satisfiability of a relation of logical consequence between a knowledge-base and a grounded theory for a closed formula $\phi$. That can be done either by querrying after learning - that is, by considering only the grounded theories that maximize the satisfiability level - or by searching a counter-example to the refered logical consequence.

Since the level of satisfiability of a symbolically generated set of formulas that belong to the axiom to be checked on a LTN gives a measure of the reasoning capability of a neural network,  the reasoning capability of a LTN network is obtained by calculating the grounding of any formula whose predicates are already grounded in the NN or by defining a new predicate in terms of existing predicates~\cite{wagner2022neural}. Also, in~\cite{wagner2022neural} it is proved that the iterative provision of negative information can improve reasoning capabilities dramatically. 

As Logical Boltzmann Machines~\cite{tran2021logical} relies on the equivalence between logical Satisfabiality and network energy minimization, these systems see Reasoning as the mentioned energy minimization or an appoximation of it. Here, to Reason about the satisfability of a prepostion, the system takes a set of assigned inputs $x_b$ and must perform a search for an assingment of truth-values for $x_b$ that satisfies the formula from what the LBM is constructed. The method normally used here is the Gibbs sampling in which the process starts with a random initialisation of $x_B$, and proceeds to infer values for the hidden units hj and then the unassigned variables $x_b$ in the visible layer of the RBM, using the conditional distributions. If the number of unassigned variables is not too large such that the partition function can be calculated directly, the Reasoning process may be performed by lowering the free energy.

%\subsection{Practical Reasoning in DFL (título copiado do artigo - modificar)}

%\alexandre{
    %In \cite{wagner2022neural}, practical reasoning in LTN is considered a task similar to inference in Machine Learning, in which conventional statistical evaluation methods are considered to select the best available grounding. The level of satisfiability of a symbolically generated set of formulas that belong to the axiom to be checked on a LTN gives a measure of the reasoning capability of a neural network. In this way, the reasoning capability of a LTN network is obtained by calculating the grounding of any formula whose predicates are already grounded in the NN or by defining a new predicate in terms of existing predicates. (texto copiado!!)
    
    %In LTN, a quantified formula can perform a kind of semi-supervised learning by expanding the set of examples in an existing domain. On the other hand, it can help with transfer learning from a data-rich domain (defined by a set of predicates) to domain with insufficient data (e.g. a new predicate). In \cite{wagner2022neural} it is proved that the iterative provision of negative information can improve reasoning capabilities dramatically. 
%}

\subsection{Relationship reasoning}

The next approach is the use of relationship reasoning. This method is farly known and consists of reasoning about relationships between entities. The NS-CL~\cite{CLEVRER}, in contrast with other approaches analysed in this survey, uses visual reasoning in order to determine an object's attributes. In this system, visual attributes are implemented as neural operators, mapping the object representation into an attribute-specific embbeding space. These visual concepts belonging to the shape attribute are later represented as vectors, which are also learned along the process, in the shape embedding space. They later measure the cosine distance between these vectors in order to compute the probability that an object is a cube by using a predefined function. They classify relational concepts, for example Left and Right, between a paior of objects similarly, except that they concatenate the visual representations of these objects to form a new representation (of their relation).

%\begin{figure}
%    \centering
%    \includegraphics[width=1.0\columnwidth]{media/NS-CL_reasoning.png}
%    \caption{Neuro-Symbolic Concept Learner reasoning process}
%    \label{fig:ns-cl_reasoning}
%\end{figure}

The NS-CL~\cite{CLEVRER} learning and reasoning heavly envolves it semantic parsing. The semantic parsing module translates a natural language question into an executable program with a hiearchy based of primitive operations which are represented in a DSL. This DSL takes care of a set of fundamental operations for visual reasoning, these being filtering out objects with certain concepts, querying the attribute of an object, etc... The operation shares, of course, the same input and output interface and, as a result, can be compositionally combined to form programs of any complexity.

\subsection{Exploration of Practical Reasoning in DFL} 

    In \cite{wagner2022openclosedworld} they argue that practical reasoning in a LTN is considered a task similar to inference in traditional Machine Learning, in which conventional statistical evaluation methods are considered to select the best grounding available. The satisfiability level of a set of formulas, symbolically generated, that belong to the axiom to be checked on a LTN gives a measure of the reasoning capability of a neural network. In this way, the reasoning capability of a LTN network can be obtained by calculating the grounding of any formula whose predicates are already grounded in the NN or by defining a set of new predicates in terms of other existing predicates.

    In LTN, a quantified formula can perform a kind of semi-supervised learning by expanding the set of examples in an existing domain. On the other hand, it can help with transfer learning from a data-rich domain (defined by a set of predicates) to domain with insufficient data (e.g. a new predicate). In \cite{wagner2022openclosedworld} it is proved that the iterative provision of negative information can improve reasoning capabilities dramatically. 

%

%\end{document}
%\orcid{0000-0002-8717-6097}

%\documentclass[acmsmall]{acmart}
%\begin{document}

\section{Explainability and Trustworthiness in Neurosymbolic Systems}
\label{s.explain}

One of the biggest criticisms of Deep Learning techniques is the lack of explainability or the ''BlackBox effect''. Most of the current state-of-the-art models are built in such a way that they are mostly opaque: the user/programmer only may gather knowledge about input-output pairing not knowing, for example, how a given feature impacts a classification decision. Even more problematic: the ubiquitous presence of AI systems in society raises questions about its negative impacts on people's lives, how much ''trustworthy'' is a system, and how we can increase it to have systems that took good, fair decisions.

Many attempts have been made for solving the ''BlackBox effect'' based on rules extraction methods~\cite{blackbox1,blackbox2,blackbox3}. The main goal of these attempts was to - based on accuracy, fidelity, consistency, and comprehensibility - search for logic rules from a given trained network~\cite{blackbox1}. There were successful attempts to implement these ideas~\cite{blackbox4}. However, those approaches were combinatorial and do not scale well when dealing with the dimensions of current neural networks (deep learning) resulting in the idea beginning to cool off, until recently when the idea of neurosymbolic systems reemerged as a combination of a global and local approach.

LTN systems~\cite{Badreddine_2022} approach to this is by increasing user and ML model communication through querying and answering. Here there are three types of queries: truth queries, value queries, and generalization truth queries.

One of the key features of neurosymbolic AI is its explainability. Given its connectionist network nature, a neurosymbolic system can be represented in a set of readable expressions. Natural language generation can be mentioned as an approach in which a user couples a deep network with sequence models to extract natural language knowledge. As can be seen in~\cite{learning_structural_emberddings} one can, instead of querying parameters of a trained model, extract relational knowledge where, by performing inference of a trained embedding network on a text data, predicates are obtained.

Another area of interest in neurosymbolic AI is program synthesis. The use of neurosymbolic AI has been proposed to build computer programs on an incremental approach based on large amounts of input-output samples~\cite{neurosymbolic_programsynthesis}. This is achieved by employing a neural network to represent partial trees. In a domain-specific language, there are tree nodes, symbols, and rules represented as vectors. It can, of course, achieve explainability through the tree-based structure of this here network.

Although nothing about explainability is directly mentioned in LBM \cite{tran2021logical},  a Knowledge Extraction may be performed on the system by querying it with appropriated unassigned variables.

Following up on the need to define queries for ML explainability,  in~\cite{arenas2021foundations} the authors propose the development of a declarative language to specify explainability queries, named FOIL. The computational complexity of FOIL queries over classes of models such as decision trees and general decision diagrams was studied, with the conclusion of this experiment being that the tractability of the FOIL evaluation was achievable by restricting the structure of the models or the portion of the FOIL language under study, deemed that such a language as proposed could be used in practice.

%\end{document}
%\orcid{0000-0002-8717-6097}

%\documentclass[acmsmall]{acmart}
%\begin{document}

\section{Applications of neuro-symbolic Systems}
\label{s.applications}

Besides the main characteristics of the neuro-symbolic Systems, we argue that is also important to highlight applications using them. In the LTN main paper \cite{Badreddine_2022} the authors show some toy-problems applications with their system. Between what is shown, there are some interesting applications like multilabel classification, semi-supervised pattern recognition, learning embeddings, and reasoning. 

\cite{wagner2021neural} shows an application of an LTN for fairness in AI. Here, the authors use First Order Logic (FOL) to input fairness constraints through a group of axioms that imputes the same treatment both to a protected and unprotected group. They discuss the results regarding the SHAP values \cite{vstrumbelj2014explaining, shapley1953value} on the impact of four observable values on the Demographic Parity metric. Then, in a second experiment, they show how LTN systems perform better regarding accuracy even with some fairness constraints on three popular datasets.

Those datasets were the Adult dataset~\cite{Dua:2019} for predicting income, the German~\cite{Dua:2019} for credit risk, and the COMPAS~\cite{larson_angwin_kirchner_mattu_2016} for recidivism. They used the same experimental setup as \cite{fairness_applications_neural} but with tuning simplifications. A comparison is proposed between their results, the one proposed in \cite{fairness_applications_neural} - another neural network-based approach that applies fairness constraints into the loss function by mean of the Lagrange multipliers - and the one proposed in \cite{fairness_applications_bayesian} - a naive-Bayes classifier -. They use gender as the protected feature in both the Adult and German datasets and race in the COMPAS~\cite{larson_angwin_kirchner_mattu_2016} dataset. Their next step was to divide our data into five arbitrary subgroups for querying, these subgroups are not to be confused with the protected and unprotected groups.

First, the network is trained without fairness constraints and they proceed to query the network to return the truth value of the LTN predicate used for classification. The output of this query helps to determine, as a proxy of similarity, their fairness constraints. Those fairness constraints were logical axioms, following LTN philosophy, it being "For all data in the subgroup equivalent result should derive from our classifier with no difference if our input is in the protected or unprotected group". With these logical axioms, one could measure how well our trained model satisfies them, giving us a better understanding of how our model perpetuates biases.

They then proceeded to use the aforementioned SHAP values to observe the demographic disparity between the protected and unprotected sections in those sub-groups. The demographic disparity is, in summation, a form of measuring if our protected and unprotected groups are receiving the same outcome from our model at equal proportions. With demographic disparity, one could measure if our algorithm, for example, benefits men over women when deciding if one should get a loan. 

\begin{figure}
  \includegraphics[width=\linewidth]{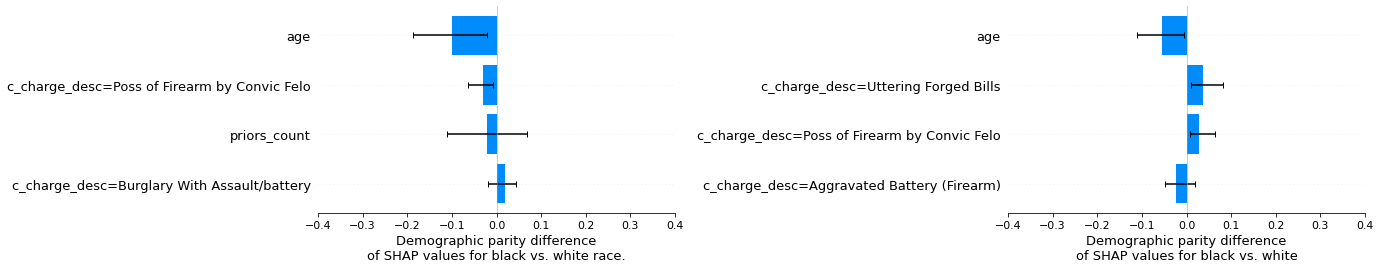}
  \caption{Disparity impact extracted with SHAP before and after LTN learning of fairness constraints in real-world data. Extracted from \cite{wagner2021neural}.}
  \label{fig:DDP}
\end{figure}

In their paper~\cite{wagner2021neural}, they use a classical example of the COMPAS~\cite{larson_angwin_kirchner_mattu_2016} dataset. They compare the demographic disparity by calculating their SHAP values. Without anti-bias axioms, the trained model exhibits, as can be seen in figure~\ref{fig:DDP}, a relevant disparity between the protected and unprotected groups. Finally, they applied fairness constraints using the same axiom mentioned before by incorporating it into the network training axioms. This resulted in a significant reduction in demographic disparity as can also be seen in figure~\ref{fig:DDP}.

The NS-CL was applied, as stated earlier, in the CLEVR~\cite{CLEVR} dataset. They claim that by applying their system to this dataset the NS-CL learns visual concepts with remarkable accuracy, allows data-efficient visual reasoning, and generalizes well to new attributes, visual composition, and language domains. 
 
They introduce a video reasoning benchmark for the complex temporal and causal structure behind the interacting objects, drawing inspiration from developmental psychology. The limitations of various current visual reasoning models on the benchmark are evaluated too.

The CLEVR~\cite{CLEVR} dataset consists of videos of shapes colliding or static. The authors use their framework (NS-CL) to answer questions concerning which shapes are present, which color they are, if they collide or not, etc... The article shows that this neuro-symbolic approach outperforms all other state-of-the-art methods by a considerable margin.

When comparing models in the generalizing to new visual compositions the NS-CL outperforms all baseline models that do not use annotations and achieves comparable results with models trained by full program annotations. Comparing now in the generalizing to new visual concepts the NS-CL also outperforms the baseline models.

The authors then extend their testing into other program domains by conducting experiments on MS-COCO~\ref{coco} images. Their models do not outperform baseline models but show competitive performance on QA (Question Answering) accuracy. Beyond question answering the authors conclude that the NS-CL effectively learns visual concepts from data.

Another possible application of neurosymbolic Computing is using it in combination with Graph Neural Networks (GNN), as proposed by~\cite{GNN}. Attention is brought to the fact that the authors only suggested this system in theory and argue that such a system, to the best of their knowledge, does not exist yet. The authors bring up attention as an example of how these two concepts can be connected. The core building blocks of GNN are graph convolution operation, which enables one to perform learning over graph inputs. As stated in~\cite{AttMech} graph convolutions, with slight variations, can be seen as an attention mechanism. Neurosymbolic systems, such as Pointer Networks~\cite{PN} implement attention mechanisms over their inputs.

The author further exercises this idea with an example related to biology and neurosymbolic Computing. Graphs are natural representations of proteins, among other molecules. In~\cite{moleculeGNN} the authors have conceived the first IA-created antibiotic ("halcin") by training a GNN to~\textit{"predict the probability that a given input molecule has a growth inhibition effect on the bacterium~\textnormal{E. coli} and using it to rank randomly-generated molecules"}~\cite{GNN}. The use of symbolic knowledge and restrictions could surely decrease the number of combinations as simulated scenarios in this context.

Neuro-symbolic AI, more precisely LTNs, can also be used for bearing fault diagnostics with constant shaft speed~\cite{bearing_faults}. The authors incorporated knowledge into deep learning in this case by proposing a new scoring function based on physical knowledge for classifying bearing faults. This scoring function later acts as domain expertise for the LTN loss function which would be later injected into the deep learning model. The authors show that their approach outperforms the baseline methods in test accuracy. 

The authors used weighted axioms to input knowledge into a neural network (NN) classifier. The parameters of the NN are optimized to maximize satisfiability. In the article, they also incorporate knowledge with a method other than LTNs Real Logic by extending the feature space with inputs describing this knowledge as in~\cite{knowledge_byfeatures}. 

Regarding experiments the authors used a pure knowledge-based classification, they did this by diving the number of correctly identified labels of a class through the number of all labels - these labels being identified using knowledge-based axioms - of a class. Real Logic in the loss function was also used, they evaluated the performance of LTNs, given the proposed knowledge base, against pure Deep Learning (DL) approaches with Multilayer Perceptrons (MLP) or Convolutional Neural Networks (CNN). The study concludes that despite little performance enhancement when looking at isolated models, a combination of neuro-symbolic and deep learning approaches does result in an accuracy increase. In summation, the articles conclude that inducting knowledge increased performance, particularly with fewer data available. They also conclude that a neurosymbolic approach helps with getting a better understanding of the dataset. The counterpoint to this is that, for them, performance gains of LTNs are not completely intuitive.

\begin{figure}
  \includegraphics[width=0.8\linewidth]{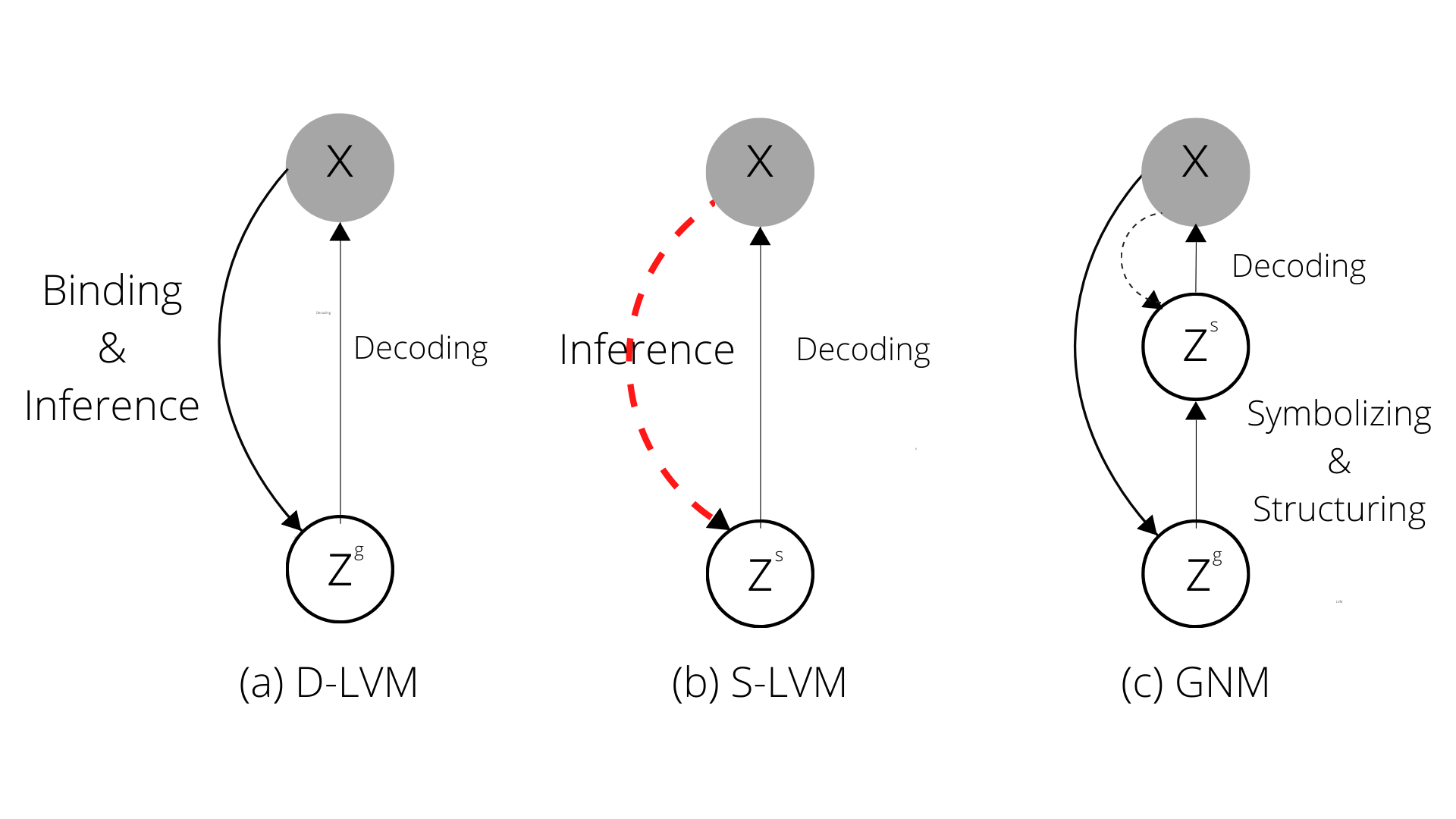}
  \caption{Graphical models of D-LVM, S-LVM, and GNM. $z^g$ is the global distributed latent representation, $z^s$ is the symbolic structured representation, and x is an observation. Adapted from \cite{jiang2020generative}.}
  \label{fig:GNN_1}
\end{figure}

\cite{jiang2020generative} show in their work the Generative neurosymbolic Machines (GNM), a probabilistic generative model, an approach to generate images based on both statistical proprieties inferred over the images' spatial distributions and explicitly structured entity-based representations. They achieve this in GNM via a two-layer latent hierarchy: the top layer generates the global distributed latent representation for flexible density modeling and the bottom layer yields from the global latent the latent structure map for entity-based and symbolic representations. Figure~\ref{fig:GNN_1} illustrates the difference between GNN, Distributed latent variable models (D-LVM), and symbolic latent variable models (S-LVM).

The authors evaluate the quality and properties of the generated images on three datasets (ARROW~\cite{apache_2019}, MNIST-4~\cite{deng2012mnist}, and MNIST-10~\cite{deng2012mnist}) in terms of clarity and scene structure, key factors on the dataset, and the model itself by using three metrics: 

\begin{itemize}
  \item Scene structure accuracy (S-Acc): they manually classified the 250 generated images per model into success or failure based on the correctness of the scene structure in the image without considering generation quality
  \item Discriminability score (D-Steps), they measure how difficult it is for a binary classifier to discriminate the generated images from the real images. Here, they considered the number of training steps required for the binary classifier to reach 90\% classification accuracy.
  \item Log-likelihood (LL) using importance sampling with 100 posterior samples
\end{itemize}

\begin{figure}
  \includegraphics[width=\linewidth]{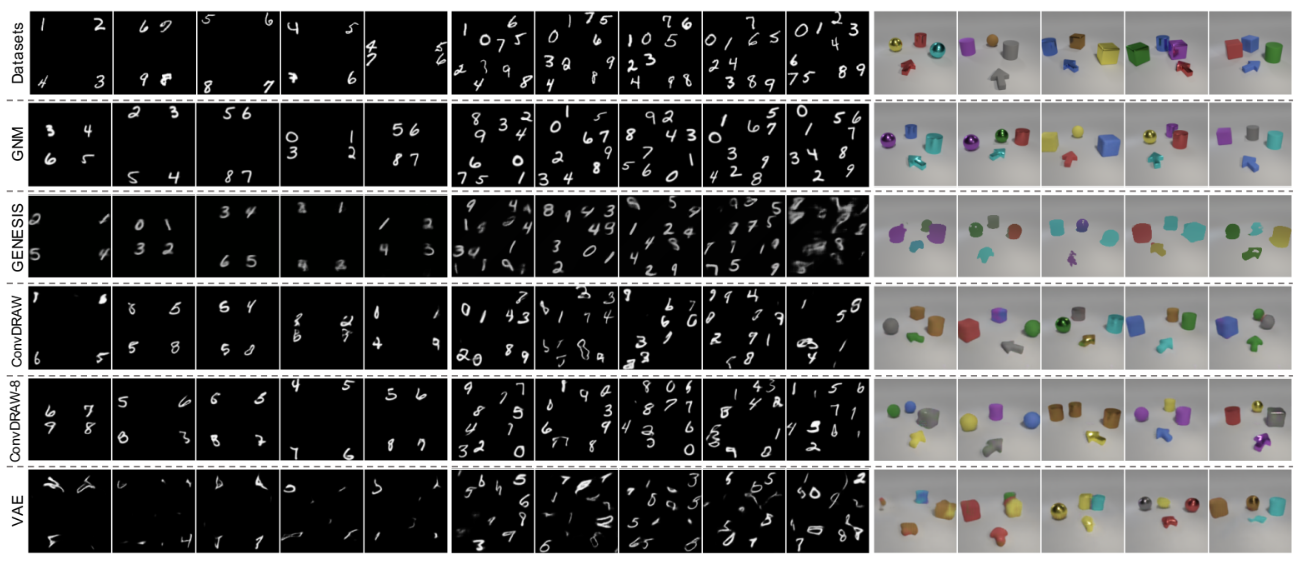}
  \caption{Datasets and generation examples. MNIST-4 (left), MNIST-10 (middle), and Arrow room (right). Extracted from \cite{jiang2020generative}.}
  \label{fig:GNN_2}
\end{figure}

The model achieves almost perfect accuracy for ARROW~\cite{apache_2019} and MNIST-4~\cite{deng2012mnist} (some generated images may be seen in figure~\ref{fig:GNN_2}) and GNM samples are significantly more difficult for a discriminator to distinguish from the real images than those generated by the baselines. Also, GNMs can be used to generate novel images by controlling an object’s structured representation, such as the position, independently of other components or by traversing the globally distributed representation and generating images, which also makes the generated objects reflect the correlation between components.

LTNs are also used in~\cite{SemanticImageInterpretation} for Semantic Image Interpretation in two important tasks: the classification of bounding boxes, and the detection of the part-of relation between any two bounding boxes. Both tasks are evaluated using the PASCAL-PART dataset~\cite{chen2014detect}.
In~\cite{SituatedLanguageUnderstanding} a model is created allowing reincorporating some ideas of classic AI into a framework of neurosymbolic intelligence using multimodal contextual modeling of interactive situations, events, and object properties. They discuss how situated grounding provides diverse data and multiple levels of modeling for a variety of AI learning challenges, including learning how to interact with object affordances, learning semantics for novel structures and configurations, and transferring such learned knowledge to new objects and situations.

Another noteworthy application of a neurosymbolic framework is the LOA (Logical Optimal Actions)~\cite{LOA_1} an architecture of action decision-making based on reinforcement learning (RL) and neurosymbolic AI. It uses LNNs (Logic Neural Networks) to bridge natural language and interaction with a set of games. 

\begin{figure}
  \includegraphics[width=0.8\linewidth]{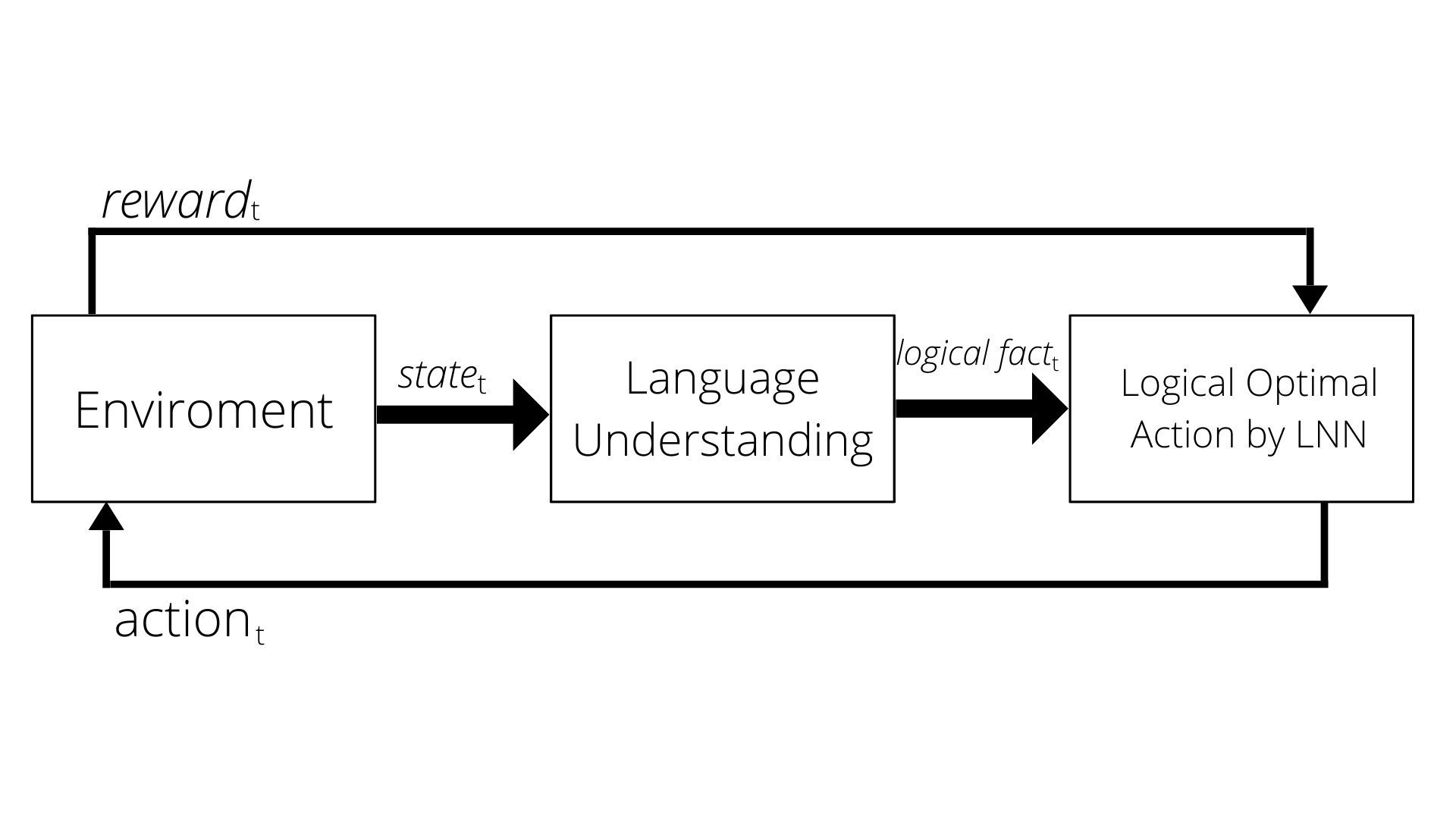}
  \caption{An architecture overview for LOA.}
  \label{fig:LNN_1}
\end{figure}

As we discussed earlier, LNNs can train the constraints and rules with logical functions in a neural network and, because every neuron in the network has a component for  a formula weighted real valued logic, it is able to compute the probability and contradiction loss for each given proposition. As a neuro-symbolic framework LNNs also follow symbolic rules, meaning that they yield interpratable and explanable representations. Figure~\ref{fig:LNN_1} shows an overview architecture for LOA.

Using those tools the authors propose a neuro-symbolic Reinforcement Learning (RL) method that uses external knowledge given prior in logical networks. They use a text-based game in order to test their new framework. The text-base game learning environment used is called TextWorld~\cite{LOA_2}. They use it as a small-scale example of a natural language-based interactive environment. The demonstration provided the authors is a web-based user interface for visualizing the game's interaction with an display for the natural text observation from the environment, typing the action sentence and showing the reward value from the each taken action. The LOA in this demonstration also takes advantage of visualizing trained and pre-defined logical rules in the LNN via the same interface. The authors argue that this will help the human user understand the benefits of introducing the logical rules via neuro-symbolic frameworks. The LOA model initially receives logic state value as logical fact from the language understanding component. This component, by turn, receives raw natural language state value from the environment. The next step for the model is the LNN. In order for the input to get optimal action for it, the action goes into the environment to execute the commanded action and is later given an reward. This reward is inputed to the LOA agent which will be trained by the action decision network (LNN) using the acquired reward value and chosen action from the network. 

The LOA demonstration proposed by the authors supports two functionalities: playing the text-based game through human interactions and visualizing the trained and pre-defined LNN to increase its interpretability for rule-acquiring. We can choose the game from some existing text-based interaction games such as the TextWorld Coin-Collector~\cite{LOA_2}. In these types of games a human player can input any action by natural language then the demonstration framework displays the raw observation output from the given environment. %Figure \ref{fig:LOA_2} shows the view for playing said game and figure \ref{fig:LOA_3} shows the view for another available game (cleanup task). 

The LNN contains simple rules for the TextWorld Coin-Collector game. For example, the rule is that the player should take the 'go west' action, after he finds the east room ("found west" then "go west") in case of a more complex LNN that supposes no repetition. As the authors themselves put: \emph{"The round box explains the proposition from the given observation inputs, the circle with a logical function means a logical function node of LNN, and the rectangle box explains an action candidate for the agent. The highlighted nodes (red node) have ‘true’ value, and non-highlighted nodes (white node) have ‘false’ value. [...], the agent found the north exit from the given observation (“Observation (t=1)”) by using semantic parser 2, then the going north room action (“go north”) is activated. [...], if the user clicks the selectable box, the LOA recommends only one action which is ‘go north’..."}

After the LNN selected the "go north" action, the player found two doors: east and south doors. However, the south door is connected to the previous room because the player took going north as an action in the previous step. In this case, the authors are using a simple LNN, thus the "go south" action is also recommended. However, the complex LNN, which has the functionality of avoiding already visited rooms, happens due to the contradiction loss in this LNN setting. This is one of the benefits of introducing neurosymbolic frameworks that human users can easily understand. The authors show in this demonstration that introducing the LNN into an RL agent brings significant benefits such as converging faster than other nonsymbolic and neurosymbolic methods while also adding more explainability to the process.

For yet another application of neurosymbolic frameworks, we can highlight the LNN-EL which is a neurosymbolic approach to short-text entity linking using Logic Neural Networks (LNN-EL)~\cite{LNN-EL}. In this system, the authors use LNNs to perform short-text entity linking. Entity linking (EL) is the task related to disambiguating textual mentions by linking them and canonical entities provided by a knowledge graph (KG) such as DBpedia. The particular type of EL discussed in this article (short text) has attracted attention due to its relevance for downstream applications such as question-answering, conversational systems, etc... The authors argue that short-text EL is particularly challenging because of the limited context surrounding mentions, thus resulting in greater ambiguity.

\begin{figure}
  \centering
  \includegraphics[width=\linewidth]{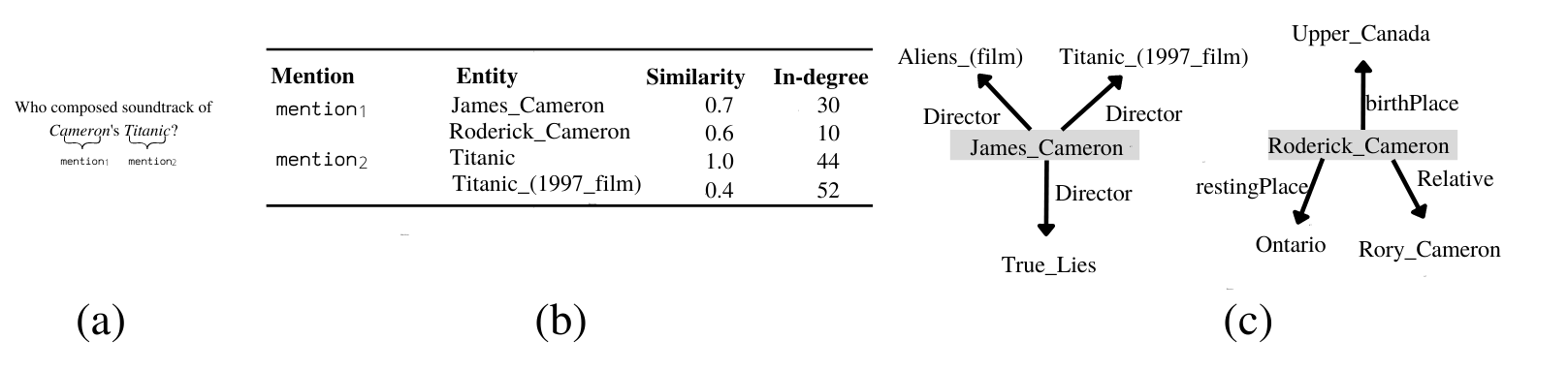}
  \caption{(a) Question with 2 mentions that shall be disambiguated against DBpedia. (b) For each mentioned candidate entity pair, the character-level Jaccard similarity along with the in-degree of the entity in the KG is shown. (c) (Partial) Ego networks for entities RoderickCameron and JamesCameron. Adapted from \cite{LNN-EL}.} 
  \label{fig:LNN_EL_1}
\end{figure}

The authors exemplify the task of short-text entity linking with the question in Figure~\ref{fig:LNN_EL_1}. The question contains $mention_1$ ($Cameron$) and $mention_2$($Titanic$) but DBpedia contains several person entities whose last name matches $Cameron$ as can be seen in Figure~\ref{fig:LNN_EL_1}(b). On one hand, given the higher in-degree - which is a result of using the more popular candidate entity in the KG, one can link $mention_1$, correctly, to $James$ $Cameron$. In terms of $mention_2$ on the other hand the correct entry is $Titanic$$(1997$$film)$ as opposed to $Titanic$ the ship which has higher string similarity. To link to the correct entity, one needs to exploit the fact that $James$ $Cameron$ has an edge connection with $Titanic$$(1997$$film)$ in the KG as shown in Figure~\ref{fig:LNN_EL_1}(c). This example is meant to provide intuition as to how priors, local features (string similarity), and collective entity linking can be exploited to overcome the limited context in short-text EL.

\begin{figure}

  \includegraphics[width=\linewidth]{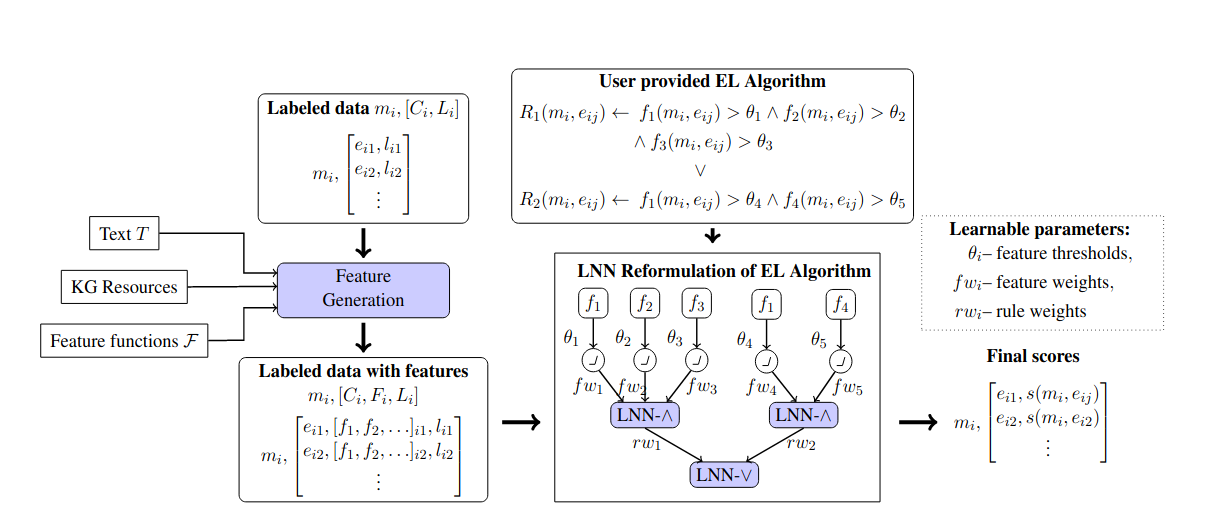}

  \caption{Overview of the authors' approach. Extracted from \cite{LNN-EL}.}

  \label{fig:LNN_EL_2}

\end{figure}

Figure~\ref{fig:LNN_EL_2} shows an overview of the authors' neurosymbolic approach for short-text EL. Given the input text T, together with the labeled data in the form ($m_i$,$C_i$,$L_i$), where $m_i$ is a mention in T, $C_i$ is a list of candidate entities $e_ij$ for the mention $m_i$ and where each $l_ij$ denotes a link or not-link label for the pair ($m_i$,$e_ij$). The authors first generate a set $F_ij$ of features for each pair ($m_i$,$e_ij$). After the feature generation, the newly labeled data with features is inputted into the LNN.

In this neurosymbolic approach to short-text EL, the EL rules are a restricted form of Fist Order Logic comprising a set of Boolean predicates connected via logical operators. The authors use the LNN logical approach to evaluate if a given mention is related to an entity. Rules can be disjuncted together to form a larger EL algorithm, thus creating a function (e.g.: Links($m_i$,e$e_ij$)). By using FOL the authors create an EL algorithm that can be easily understood and manipulated by users. However, the authors highlight that to obtain competitive performance against the best deep learning approaches, this system requires a significant amount of manual effort to fine-tune certain aspects.

Although the pure LNN model underperforms compared to black-box deep learning methods it is competitive and outperforms other logic-based approaches. Furthermore, LNN-EL$ens$, which combines the core LNN-EL with deep learning approaches, easily beats those same deep learning approaches in almost all datasets. The authors argue that the leverage this model has over other models, these being deep learning or other techniques, is its explainability, interpretability, and transferability.

They test the model transferability by training the model on one dataset and evaluating it on the other two datasets. They observe that the model transfers reasonably well even if the training is done on a very small dataset. Benchmark deep learning models outperform the LNN-EL in transferability but the LNN-EL needs significantly less data for training to achieve reasonable performance. 

Our last application also makes use of LNNs. In this article,~\cite{KBQA} the authors use LNNs as a reasoner for knowledge base question answering (KBQA). Knowledge base question answering - a sub-field of question answering as a whole - is a very important task in natural language processing (NLP). KBQA demands that a system answer natural language questions based on facts available in a knowledge base (KB). This system retrieves facts from a KB through structured queries, these queries often contain multiple triples that represent the steps or antecedents required for obtaining the answer, thus enabling a transparent and self-explanatory form of QA. 

In the field of KBQA, neural networks are the most common approach to this type of problem. The authors propose neuro-symbolic Question Answering (NSQA), a modular knowledge base question answering system that would: a) delegate the complexity of understanding natural language questions to AMR (Abstract Meaning Representation) parsers~\cite{AMR_1,AMR_2} - which is one of the benchmark parsers types for semantic representation; b) reduce the need for end-to-end training data using a pipeline architecture where each module is trained for its specific task; c) facilitate the use of an independent reasoner via an intermediate logic form (LNNs). 

Suppose that a question, given in natural language, is input into the system; it firstly parsers the question into an abstract meaning representation (AMR) graph, then transforms this graph into a set of candidate KB-aligned logical queries; the system then uses an LNN to reason over KB facts, producing answers to KB-aligned logical queries. In this survey, the actual parsing or the AMR to KG Logic process is not explored. However, it focuses on the LNN reasoner. 

Given the KB-aligned logical queries, the NSQA uses LNNs as a first-order logic neuro-symbolic reasoner. The system uses LNNs type-based reasoning to eliminate queries based on inconsistencies making use of the type hierarchy in the KB. The system also uses the LNNs geographic reasoning to answer questions such as "Was Cartola born in Salvador?", this happens because the entities related to $dbo:birthplace$ are generally cities, but the question requires a comparison of countries. The authors address this manually by adding logic axioms to perform the required transitive reasoning for $dbo:birthplace$.

For their experimental evaluation, the authors use 4 datasets. They first evaluate NSQA against four systems, two being graph-driven approaches, one being a KB agnostic approach and the last being an ensemble of entity and relation linking modules and train a Tree-LTSM model for query ranking. The NSQA system outperforms three of the four systems in two of the four datasets. 

This survey shall highlight the section in which the authors analyze the LNN reasoner results.

The authors evaluate the performance of NSQAs under a LNN reasoner and a deterministic translation of query graphs to SPARQL. They tested it in one dataset and the LNN approach outperforms the deterministic translation by a 2.9\% margin. Table \ref{tab:questions} shows the type of question the NSQA system is capable of answering. 

%\begin{landscape}
  \begin{table}[h!]
  \centering
  \resizebox{\columnwidth}{!}{%
  \begin{tabular}{lll}
  \hline
  Question Type/Reasoning & Example                                                        & Supported                 \\ \hline
  Simple                  & Who is the president of Brazil                                 & \checkmark  \\
  Multi-Relational & Give me all actors starring in movies directed by Glauber Rocha & \checkmark \\
  Count-based             & How many books did Machado de Assis write?                      & \checkmark  \\
  Superlative             & What is the highest mountain in Chile?                         & \checkmark  \\
  Comparative             & Does A Grande Familia have more episodes than Grey's Anatomy?  &                           \\
  Geographic              & Was Cartola born in Argentina?                                 & \checkmark  \\
  Temporal                & When will start the final match of the football world cup 2022 &                           \\ \hline
  \end{tabular}%
  }
  \caption{Question NSQA system is capable of answering.}
  \label{tab:questions}
  \end{table}
  %\end{landscape}

%\end{document}
%\orcid{0000-0002-8717-6097}

%\documentclass[acmsmall]{acmart}
%\begin{document}

\section{Conclusion}
\label{s.conclusion}

Neurosymbolic Artificial Intelligence is a prominent approach that combines learning over data distribution and reasoning on prior and learned knowledge. The present survey compiles the most recent works on neurosymbolic AI and their applications. Also, it provides a taxonomy of approaches on each aspect of the area, like Knowledge Representation, Learning and Reasoning, as well as a brief historical introduction.

About $14$ papers were reviewed with a focus on building systems with prior knowledge - or premises - that would significantly reduce the amount of data needed to achieve good performance concerning traditional metrics like accuracy. Table \ref{tab:comparison} shows a taxonomic comparison between the surveyed models. 

In order to further enrich the discussion and field of neurosymbolic AI we will now highlight blindspots we found concerning the articles we read. This highlighted points could be a jumping point for future work conserning neurosymbolic AI:
\begin{itemize}
  \item \textbf{Ability to save models in LTNs}: When creating a model sometimes it can takes hours or even days for it to run completely and in LTNs there is no feature to save this model for further use without re-training. 
  \item \textbf{Explanability}: Although the models try (and succeed mostly) in being explanable this explanation is mostly aimed at the scientist or specialist, thus not being user-friendly.
  \item \textbf{Comparison}: When comparing models we felt a lack of more robust comparisons between neurosymbolic approaches and non-logic/symbolic approaches such as deep learning, decision trees and so on. 
  \item \textbf{Fairness and algorithmic bias}: In our tests and read articles we came across the use of neurosymbolic AI for fairness and bias mitigation. We found out they mix very well and there is a lack of artiles exploring this relation. 
\end{itemize}

While these systems often underperform traditional, pure Deep Learning techniques, the amount of data, computational costs, and the possibility to have a better understanding and guidance over the system justify further research in this direction as an alternative to evergrowing DL models.

% \usepackage{booktabs}

% \usepackage{booktabs}
% \usepackage{graphicx}
% \usepackage{lscape}
%\begin{landscape}
    \begin{table}[]
    \centering
    \caption{Comparison between the models a}
    \label{tab:comparison}
    \resizebox{\columnwidth}{!}{%
    \begin{tabular}{@{}llllll@{}}
    \toprule
    Model &
      \begin{tabular}[c]{@{}l@{}}Knowledge \\ Representation\end{tabular} &
      Learning &
      Reasoning &
      Explainabilty &
      Applications \\ \midrule
    LTN \cite{Badreddine_2022} &
      Tensorisation &
      \begin{tabular}[c]{@{}l@{}}Vertical Hybrid\\ Learning\end{tabular} &
      \begin{tabular}[c]{@{}l@{}}Approximate \\ Sat\end{tabular} &
      \begin{tabular}[c]{@{}l@{}}Knowledge \\ Extraction\end{tabular} &
      \begin{tabular}[c]{@{}l@{}}Fairness \\ Constrains \cite{wagner2021neural}, \\ Semantic Image 
        \\ Interpretation \cite{SemanticImageInterpretation} ,\\ Situated Language \\ Understanding 
        \cite{SituatedLanguageUnderstanding}, \\Neural-Symbolic \\Reasoning\cite{wagner2022neural}\end{tabular} \\ \midrule
    NLM \cite{dong2019neural} &
      Tensorisation &
      \begin{tabular}[c]{@{}l@{}}Inductive Logic \\ Programming\end{tabular} &
      \begin{tabular}[c]{@{}l@{}}Forward \\ Chaining\end{tabular} &
      - &
      \begin{tabular}[c]{@{}l@{}}Family Tree \\ Reasoning,\\ Block World, sorting,\\ path fiding  \cite{dong2019neural}\end{tabular}\\ \midrule
    LBM \cite{tran2021logical} &
      Propositionalization &
      \begin{tabular}[c]{@{}l@{}}Inductive Logic \\ Programming\end{tabular} &
      \begin{tabular}[c]{@{}l@{}}Approximate \\ Sat\end{tabular} &
      \begin{tabular}[c]{@{}l@{}}Knowledge \\ Extraction\end{tabular} &
      \begin{tabular}[c]{@{}l@{}}Satisfying set \\ search,\\ Classification \cite{tran2021logical} \end{tabular} \\ \midrule
    LNN \cite{LNN} &
      Tensorisation &
      \begin{tabular}[c]{@{}l@{}}Vertical Hybrid\\ Learning\end{tabular} &
      \begin{tabular}[c]{@{}l@{}}Forward/Backwards\\ Chaining\end{tabular} &
      \begin{tabular}[c]{@{}l@{}}Knowledge \\ Extraction\end{tabular} &
      \begin{tabular}[c]{@{}l@{}}LOA \cite{LOA_1}, \\ EL \cite{LNN-EL}, KBQA \cite{KBQA}\end{tabular} \\ \midrule
    NS-CL \cite{CLEVRER} &
      Propositionalization &
      \begin{tabular}[c]{@{}l@{}}Vertical Hybrid\\ Learning\end{tabular} &
      \begin{tabular}[c]{@{}l@{}}Relationship \\ Reasoning\end{tabular} &
      - &
      CLEVRER \cite{CLEVRER}\\ \bottomrule
    GNM \cite{jiang2020generative} &
      Tensorisation &
      \begin{tabular}[c]{@{}l@{}}Evidence \\Lower Bound\end{tabular} &
      \begin{tabular}[c]{@{}l@{}}Approximate \\ Sat\end{tabular} &
      - &
      Image Generation \cite{jiang2020generative}\\ \bottomrule
    \end{tabular}%
    }
    \end{table}

\section*{Acknowledgments}
The authors are grateful to the Eldorado Research Institute.

%Bibliography
\bibliographystyle{unsrt}  
\bibliography{sections/references}

\end{document}